\documentclass[letterpaper, 10 pt, conference]{ieeeconf}  

\IEEEoverridecommandlockouts                              
\usepackage{cite}

\usepackage[pdftex]{graphicx}
\usepackage{tabularx}
\usepackage{multirow}

\usepackage{amsmath}
\usepackage{amssymb}
\usepackage{amsfonts}
\usepackage[ruled, vlined, linesnumbered]{algorithm2e}

\usepackage{booktabs}
\usepackage{placeins}

\usepackage{balance}

\usepackage{graphicx, subcaption}
\usepackage{float}
\usepackage[colorlinks=true,urlcolor=blue]{hyperref}
\hypersetup{
    linkcolor = {black},
    citecolor = {black}
}

\usepackage{xcolor}

\usepackage[font=small]{caption}

\usepackage{balance}

\usepackage{tikz}
\usetikzlibrary{automata, positioning, arrows}

\usepackage{todonotes}

\let\labelindent\relax
\usepackage{enumitem}

\title{Design and Aerodynamic Modeling of MetaMorpher: A Hybrid Rotary and Fixed-Wing Morphing UAV}
\author{Anja Bosak*, Dorian Erić*, Ana Milas, Stjepan Bogdan 
\thanks{*Authors contributed equally. All authors are with the University of Zagreb, Faculty of Electrical Engineering  and Computing, LARICS Laboratory for Robotics and Intelligent Control Systems, Unska 3, 10000 Zagreb, Croatia; {\tt\small (anja.bosak, dorian.eric, ana.milas, stjepan.bogdan)@fer.unizg.hr}}}

\begin{document}

\maketitle

\begin{abstract}

In this paper, we present a generalized, comprehensive nonlinear mathematical model and conceptual design for the \textit{MetaMorpher}, a metamorphic Unmanned Aerial Vehicle (UAV) designed to bridge the gap between vertical takeoff and landing agility and fixed-wing cruising efficiency. Building on the successful design of the \textit{spincopter} platform, this work introduces a simplified mechanical architecture using lightweight materials and a novel wing-folding strategy. Unlike traditional rigid-body approximations, we derive a nonlinear flight dynamics model that enables arbitrary force distributions across a segmented wing structure. This modularity allows for testing different airfoils, mass distributions, and chord lengths in a single environment. As part of this work, various flight modes were specifically tested and analyzed in the \textit{Simulink} environment. The results show that the model behaves predictably under different structural configurations, demonstrating its reliability as a tool for rapid design evaluation.

\end{abstract}

\IEEEpeerreviewmaketitle

\section{Introduction}


The adoption of Unmanned Aerial Vehicles (UAVs) has enabled integration across a wide range of civilian sectors. Today, UAVs can perform complex tasks in various environments, such as monitoring traffic, crops, and pollution, as well as inspecting critical infrastructure like power transmission lines, wind turbines, and solar power plants \cite{li2018review}. These missions often require operation in limited spaces for takeoff and landing, while also demanding high-endurance flight to cover large areas.

Metamorphic UAVs are at the forefront of aerial robotics in meeting these demands, evolving from rigid structures to bio-inspired, adaptable vehicles \cite{sui2022optimum}. By altering their morphology, including wing sweep, camber, and chord distributions, these vehicles can optimize their aerodynamic profiles for different flight modes, such as hovering, cruising or aggressive maneuvering \cite{obradovic2011modeling, cevdet2021review}. This adaptability is crucial for applications like environmental sensing, where long-range coverage is combined with stationary monitoring \cite{manfreda2018}.
Despite the clear advantages of such multimodal systems, functioning as both a hover-type vehicle and a high-efficiency gliding vehicle introduces significant nonlinearities and structural complexities. 

This paper bridges the gap between bio-inspired theory and practical implementation by presenting an integrated modeling framework and mechanical design for the metamorphic UAV, called \textit{MetaMorpher}. The MetaMorpher is an advanced vehicle capable of vertical takeoff, hovering, and landing, with the aerodynamic efficiency required for long-range cruise flight. 
It represents an advancement of our previously developed \textit{spincopter}, a UAV that demonstrated stability and maneuverability with its unique twin-wing rotating structure. The spincopter was first designed as a small-scale prototype \cite{Orsag2011} and later scaled to a larger, more robust model presented in \cite{Orsag2013}. The vehicle's dual-propeller propulsion system was successfully validated in complex real-world environments, and its effectiveness as a stable UAV for autonomous surveillance was demonstrated in \cite{Haus2013}.

Building on this proven spincopter model, the MetaMorpher extends the design to have a transforming plane-like configuration for long distance missions, where being in a hover configuration is not beneficial. To try and keep the goal of high efficiency flight, a flying wing configuration is chosen, for its efficient aerodynamic design and beneficial shape for transitioning from a spincopter design. The MetaMorpher leverages advances in lightweight materials to simplify the airframe and reduce mechanical complexity. This material evolution ensures reliable metamorphosis by combining the stability of the original spincopter with high-efficiency cruise capabilities. To harness this mechanical simplicity and optimize the transition between flight modes, a highly flexible modeling approach is required.
Thus, in this paper, we focus on the development of a generalized mathematical model, tailored to simulate both the rotary-wing and fixed-wing configurations within a single environment. This modular framework establishes a control-oriented plant model capable of rapid evaluation of various airfoils, mass distributions, and structural dimensions.
While existing research often simplifies metamorphic UAVs as rigid bodies, our approach models a UAV under arbitrary force and moment distributions. Our model divides the wing into a variable number of segments, each with its own aerodynamics.
The main contributions of this work are:
\begin{enumerate}

    \item Conceptual design of a metamorphic UAV including the mechanical architecture of the first MetaMorpher prototype, featuring a design that supports both rotary-wing and fixed-wing configurations through a wing-folding strategy.
    \item A nonlinear flight dynamics model that incorporates segmented aerodynamics to simulate different flight modes (hover and cruise), structural materials, and mass distributions, providing a robust environment for rapid design iteration before physical prototyping.
    \item The complete model developed in Simulink is available as an open-source repository to benefit the community and support the reproducibility of this research. The repository of our model can be found at\footnote{\url{https://github.com/larics/metamorph_uav_matlab}}.
\end{enumerate}

This paper is divided into five sections. Section \ref{sec:related_work} compares our work with existing related work and outlines similarities and differences. Section \ref{sec:modeling} outlines the design of the MetaMorpher, focusing specifically on wing design and airfoil selection. In Section \ref{sec:mat_model} we present the nonlinear mathematical model, while in Section \ref{sec:sim_analysis} we present the results obtained in various simulation scenarios. Finally, in Section \ref{sec:conclusion}, we give conclusions and directions for future work.

\section{Related work}
\label{sec:related_work}

Metamorphic UAVs are designed to overcome the limitations of conventional drones, which are typically restricted to a single fixed structural configuration. Morphological transformation can be achieved through various design approaches, including bio-inspired mechanisms and hybrid VTOL platforms. Although many studies focus on mechanical design and experimental validation, comprehensive mathematical models capable of simulating diverse flight modes within a single environment remain limited.

Most existing papers are either tailored to a specific hardware prototype or simplify the UAV as a rigid body. Few use software simulators to test the prototype or specific parts of the design process. For instance, \cite{Anuar2025} and \cite{Shams2021} use XFLR5 \cite{xflr5} for aerodynamic analysis, focusing on stability and lift-to-drag ratios, but lack a unified dynamic simulation environment. Others, such as \cite{Coban2023}, describe the detailed design and manufacturing process but rely on theoretical equations without specifying simulation software. Additionally, Bai et al. \cite{Bai2019} used simplified analytical approaches to evaluate the samara-inspired revolving-wing robot and the \textit{SplitFlyer Air} modular quadcopter \cite{Bai2022Splitflyer}. By employing a small-angle assumption, the complex revolving dynamics are reduced to a high-order linear system \cite{Crandall1995}. However, while such analytical simplifications are effective for assessing stability in specific flight regimes, they may not fully capture highly nonlinear transitions or the effects of arbitrary force distributions in more complex metamorphic designs.

To address these nonlinearities, some researchers have used high-fidelity simulation tools for specific flight modes. For instance, Yuksek et al. \cite{Yuksek2016} developed a complete six-degree-of-freedom nonlinear model for a tilt-rotor UAV, specifically incorporating propeller-induced airstream effects during transition. Similarly, Zhao et al. \cite{Zhao2021} used Kriging-based aerodynamic modeling to manage the control allocation of flying wing UAVs with morphing trailing edges, though their focus remained on specific control surfaces rather than global structural metamorphosis.

Nature-inspired platforms further illustrate the complexity of simulating multiple modes. Win et al. \cite{Win2021} used MATLAB Simscape Multibody to model the folding dynamics of a single-actuator rotary wing and Sufiyan et al. \cite{Sufiyan2021, Sufiyan2022} introduced a multimodal vehicle capable of rotor-wing, tailsitter and cruise modes. Notably, the work by Sufiyan et al. \cite{Sufiyan2021} employs a collaborative co-evolutionary approach to optimize both design and control parameters. However, their implementation relies on Python-based frameworks.
Furthermore, Low et al. \cite{Low2017} introduced the Transformable HOvering Rotorcraft (THOR), which is capable of both hover and cruise modes. In each flight mode, the vehicle is treated as a unified rigid body. Our work builds on these foundations by introducing a segmented wing architecture that captures localized aerodynamic effects and structural nonlinearities, which are often simplified in the early stages of hybrid UAV design \cite{Low2018}. While this platform demonstrates successful mode switching, the current state of the art tends to focus on mechanical designs and experimental testing of a single hardware iteration. Additionally, research that focuses primarily on mechanical optimization, such as the work by \cite{Woods2014Adaptive} and \cite{Yang2021} on buckling models and parallelogram mechanisms, often employs analytical models that are not integrated into a full-flight simulation.

Our work bridges these gaps by introducing a segmented wing architecture within a unified Simulink framework. We propose a modeling framework that covers hover and cruise modes and transitions. To the best of our knowledge, no existing mathematical framework enables rapid exploration of vehicle design with a specific focus on segmented aerodynamic surfaces across a few flight regimes. This modularity ensures that MetaMorpher can be used to optimize vehicle design for stability and efficiency before physical implementation or control strategy development. Finally, this mathematical model provides a foundation for future testing of metamorphic transitions and the implementation of advanced control algorithms.

\section{MetaMorpher design}
\label{sec:modeling}

\begin{table*}[t]
\centering
\caption{Nomenclature and symbol description.}
\label{tab:nomenclature}
\begin{tabular}{@{}ll ll@{}}
\toprule
\textbf{Symbol} & \textbf{Description} & \textbf{Symbol} & \textbf{Description} \\
\midrule

$\mathcal{F}_P, \mathcal{F}_S, \mathcal{F}_B, \mathcal{F}_W$ & Port, starboard, body and world frames 
& $L, D$ & Lift and drag forces \\

$\mathbf{p}$ & Position vector in $\mathcal{F}_W$ 
& $\rho$ & Air density \\

$\mathbf{v}$ & Linear velocity vector in $\mathcal{F}_B$ 
& $q$ & Dynamic pressure \\

$\boldsymbol{\omega}$ & Angular rate vector in $\mathcal{F}_B$ 
& $i$ & Segment index \\

$m_{uav}$ & Total mass of the UAV 
& $A_{P}^{i}, A_{S}^{i}$ & Segment reference area \\

$\alpha$ & Angle of attack 
& $b_{\text{seg}}$ & Segment span \\

$Re$ & Reynolds number 
& ${\mathbf{r}_{AC}}_P^i, {\mathbf{r}_{AC}}_S^i$ & Aerodynamic center vector in $\mathcal{F}_B$ \\

$c$ & Chord 
& $\mathbf{F}_{P}^i, \mathbf{F}_{S}^i, \mathbf{M}_{P}^i, \mathbf{M}_{S}^i$ & Segment force and moment \\

$c_l, c_d, c_m$ & Aerodynamic coefficients 
& ${V_{air}}_P^i, {V_{air}}_S^i$ & Air-relative velocity magnitude \\

\bottomrule
\end{tabular}
\end{table*}

The MetaMorpher is designed to fly in different modes, including hovering and cruising, enabled by a novel structural transformation. Although the specific mechanical implementation of the metamorphosis is beyond the scope of this paper, the UAV's overall architecture is developed with this functionality in mind. The design adheres to strict requirements, such as dual-mode capability without excessive mechanical complexity, a thrust-to-weight ratio greater than one for vertical ascent, and a target wing loading of $5\,\text{kg/m}^2$. The target wing loading is calculated as follows \cite{Anuar2025}:
\begin{equation} \label{eq:wing_loading}
    W_s = \frac{m_{uav}}{A},
\end{equation}

where $m_{uav}$ is the total mass of the UAV and $A$ is the total wing area. This wing loading target provides a starting point for designing the UAV. Another key parameter is the wingspan, as it determines the overall size. For the first prototype, a wingspan of 70 cm was chosen based on manufacturing capabilities. 
Considering the required electronics and mechanical parts for a UAV with this wingspan, a rough mass estimate of 450 g was made. Using Eq. \ref{eq:wing_loading}, a wing area of 900 cm² was calculated. With these initial parameters, a more in-depth analysis can be conducted.

\subsection{Wing Design and Airfoil Selection in XFLR5 Simulator}
\label{subsec:wing_design}

\begin{figure}[t]
	\centering
 \includegraphics[width=1.0\columnwidth]{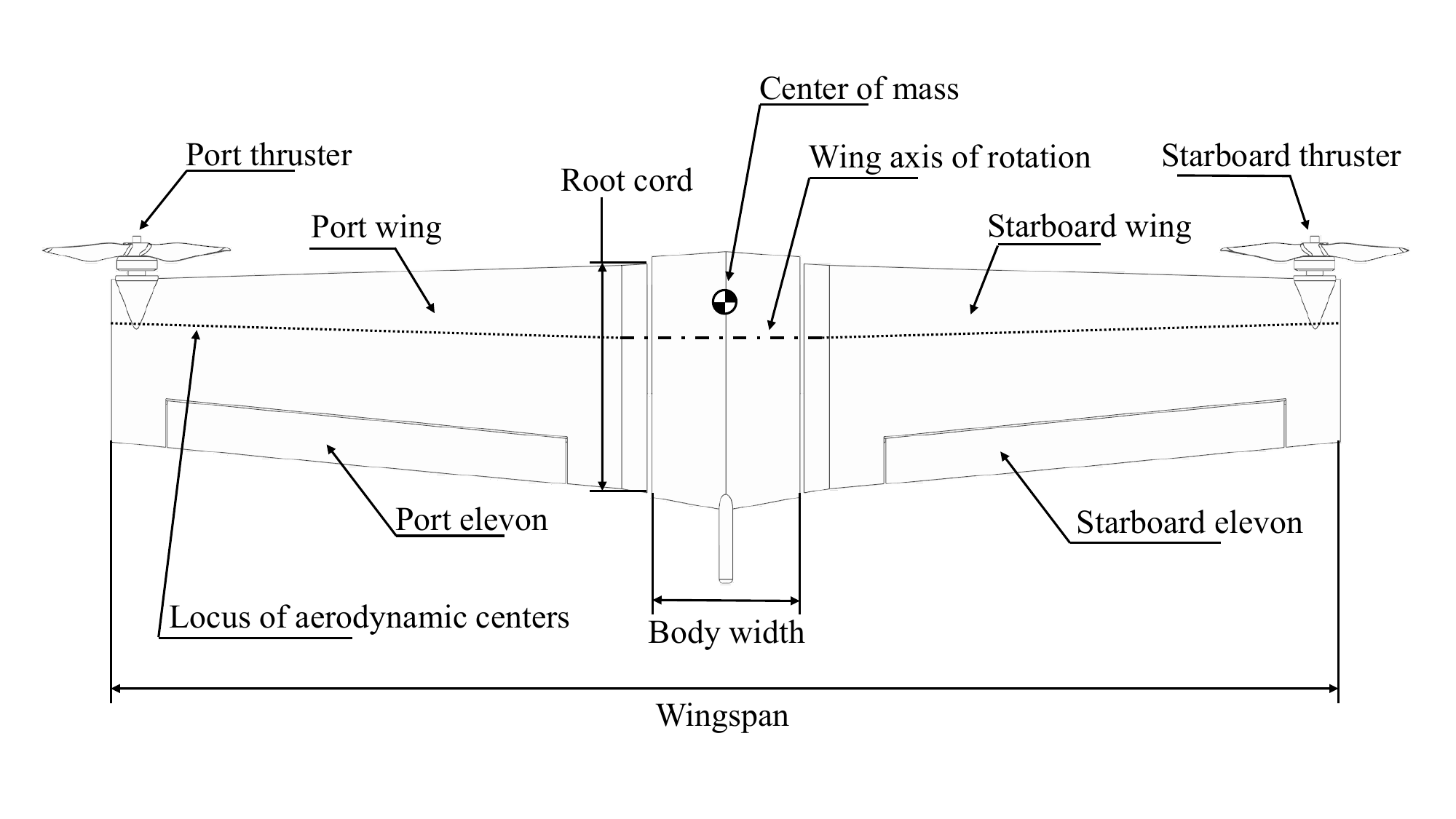}
	\caption{MetaMorpher in the XFLR5 environment, used for quasi-static stability analysis and the generation of aerodynamic polars.}
	\label{fig:UAV_parts_XFLR5}
\end{figure}

XFLR5 is a widely used open-source tool for aerodynamic analysis of airfoils, wings, and planes operating at low Reynolds numbers \cite{xflr5}. Based on the XFoil code, it uses Lifting Line Theory (LLT) and the Vortex Lattice Method (VLM) and extending on that, implements a 3D Panel Method to estimate aerodynamic coefficients and evaluate flight stability. While highly effective for rapid aerodynamic characterization and initial design iterations, as shown in studies by \cite{Shams2021} and \cite{Anuar2025}, XFLR5 is primarily a steady-state or quasi-static solver. 

To establish a general direction for wing design, inspiration was drawn from the THOR UAV \cite{Low2018}, which has a weight and wingspan similar to the MetaMorpher. A distinctive feature of the MetaMorpher is the inversion of wing profiles between operating modes. To maintain pitch stability in a tailless flying-wing configuration, the cruise mode airfoil needs to have a reflex camber design \cite{Shams2021}. Based on the comprehensive evaluation of low-Reynolds-number candidates by Shams et al. \cite{Shams2021}, and confirmed by our initial comparative analysis, the Eppler E387 reflex camber profile was selected for the cruise regime to ensure longitudinal stability. XFLR5 then calculates the lift, drag, and pitching moment coefficients, $c_l$, $c_d$, and $c_m$, respectively, which are needed for further analysis. Using multiple models with different wing characteristics, a wing was designed with a $0^\circ$ twist, no dihedral angle, no root-tip sweep, a 0.688 taper ratio, and a 160 mm root chord length.
To evaluate the effects of these characteristics on stability in cruise mode, a full-scale 3D model was developed in XFLR5 (Fig. \ref{fig:UAV_parts_XFLR5}). This model incorporates the computed mass distribution and the geometric constraints of the wing panels.

A symmetric NACA 0010 profile was selected for hover mode to match the geometric perimeter length of the E387 airfoil, ensuring a seamless structural transition during the morphing process. The need for a secondary profile arises from the wing's rotation about the wing axis (Fig. \ref{fig:UAV_parts_XFLR5}): in cruise mode, the z-axes of the body frame ($\mathcal{F}_B$), port frame ($\mathcal{F}_P$), and starboard frame ($\mathcal{F}_S$) are coaligned, whereas transitioning to hover mode rotates one wing so that its $z$-axis acquires a negative component along the $z$-axis of $\mathcal{F}_B$, effectively inverting one wing profile relative to the other. This is the main reason for a morphing wing profile, and the symmetrical wing profile was chosen for manufacturing reasons. The 2D XFLR5 simulations for the two selected profiles generate aerodynamic coefficients across a comprehensive range of Reynolds numbers ($Re$) and angles of attack ($\alpha$), which are then exported.

The data was exported as a lookup table and used in Simulink to calculate forces and moments on the wing segments in real time. This allows the simulation to capture transient effects during dynamic flight, which are beyond the scope of steady-state solvers like XFLR5.
A list of symbols and their respective description used in this paper is summarized in Table \ref{tab:nomenclature}.

\section{Nonlinear Mathematical Model}
\label{sec:mat_model}

\begin{figure}[t]
	\centering
 \includegraphics[width=1.0\columnwidth]{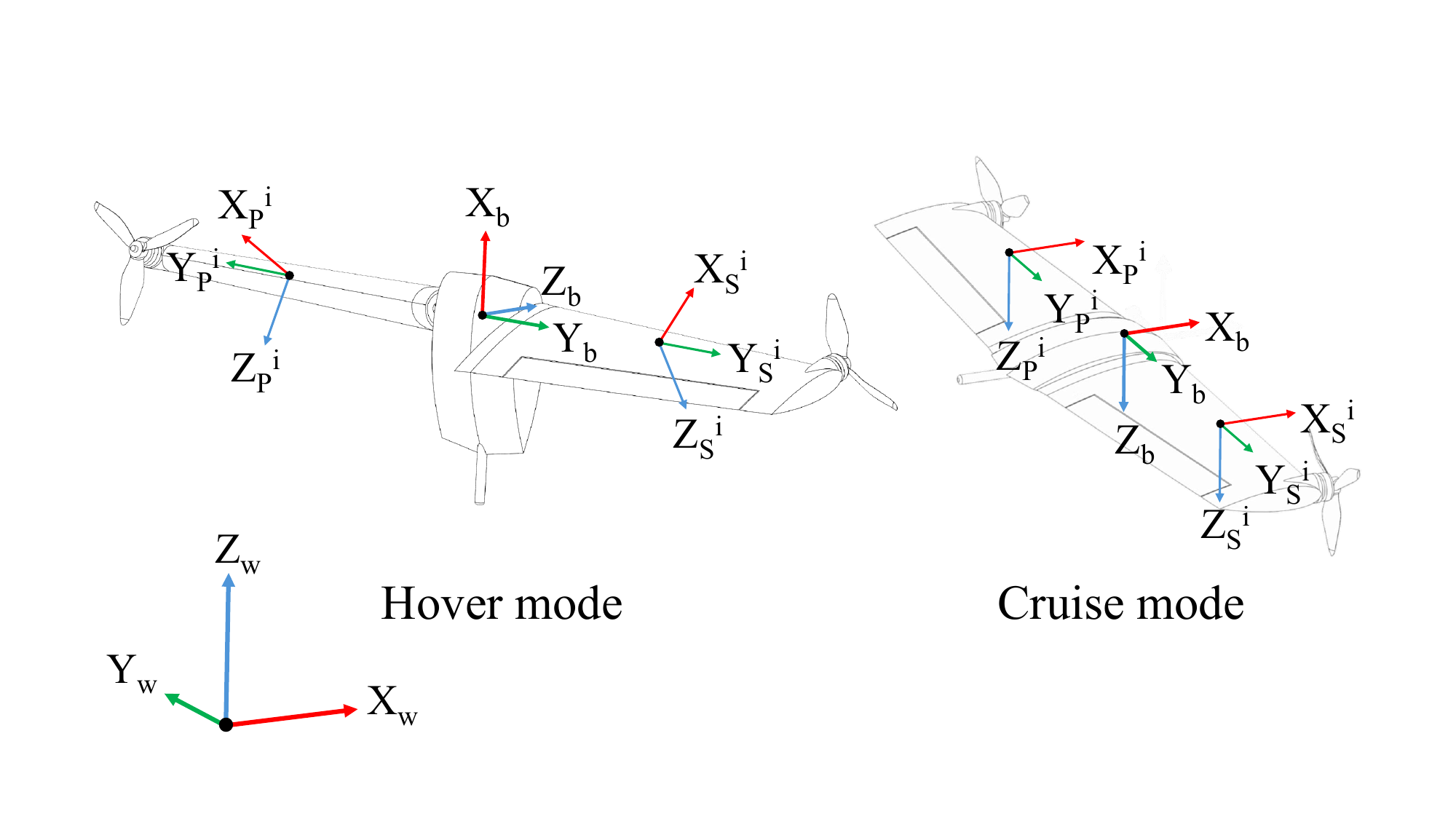}
	\caption{Reference frames for the hover and cruise mode with the body frame ($\mathcal{F}_B$) and the frames of the $i$-th segment of both the port and starboard wings.}
	\label{fig:coordinate_systems}
\end{figure}

\begin{figure}[t]
    \centering
    \includegraphics[width=1.0\linewidth]{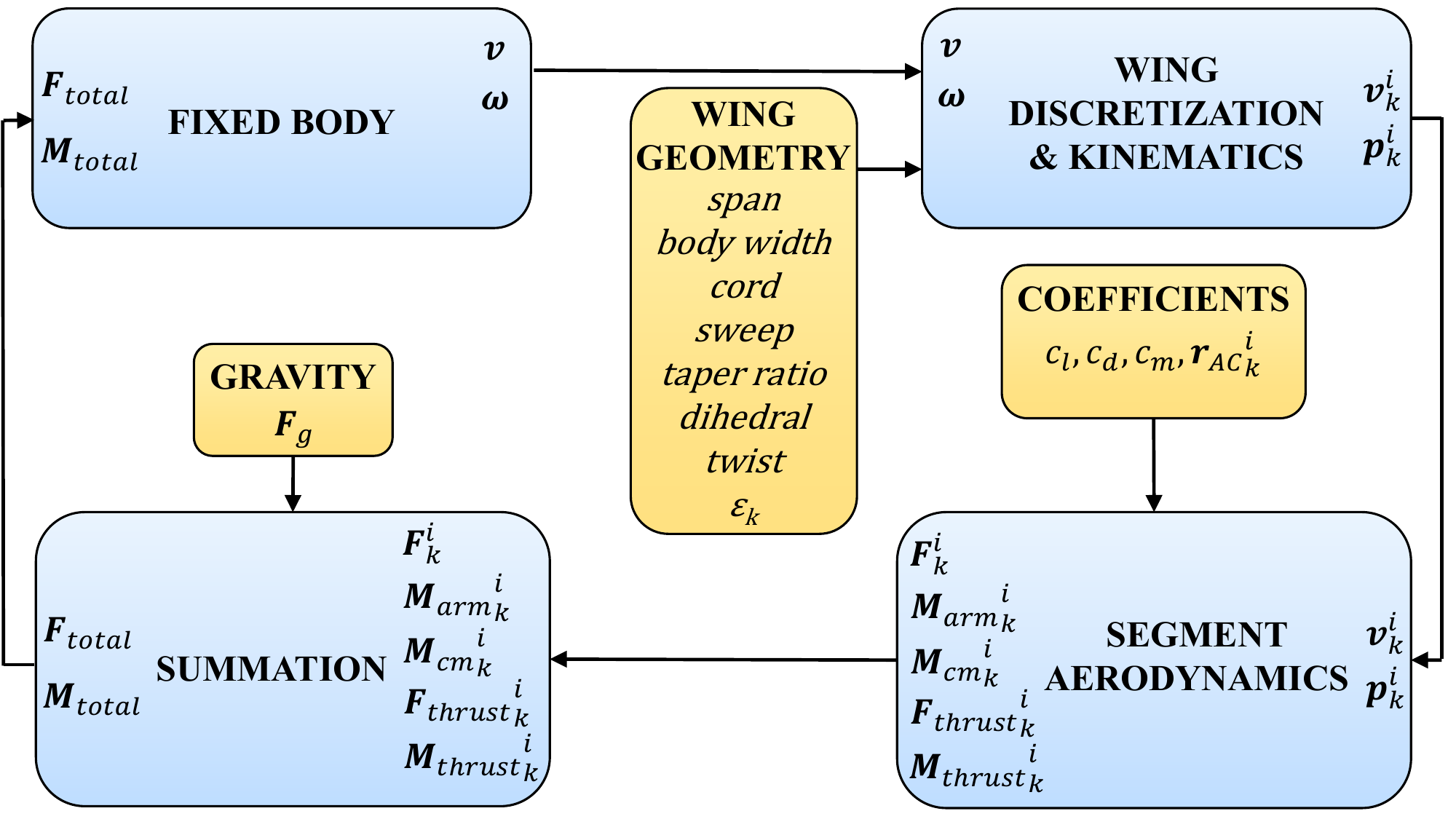}
    \caption{Block diagram of the nonlinear aerodynamic model implemented in Simulink.}
    \label{fig:model_structure}
\end{figure}
In this section, we present a nonlinear six-degree-of-freedom (6-DoF) mathematical model developed to simulate the MetaMorpher’s flight dynamics. Unlike standard rigid-body models, this framework accounts for segmented aerodynamic forces that occur during metamorphosis.
The coordinate systems used to derive the equations of motion are shown in Fig. \ref{fig:coordinate_systems}.

The nonlinear dynamic model of the vehicle was implemented in Simulink using selected components from the toolbox called Aerospace Blockset. The model is shown in Fig. \ref{fig:model_structure}. It is composed of four primary modules: rigid-body dynamics, wing geometry and segmentation, segment aerodynamics, and force and moment summation. Each component is described in detail in the following subsections.

 \subsection{Fixed Body}
The core of the model consists of a 6-DoF rigid-body block representing a fixed-body aircraft. 
All equations of motion are expressed in the body-fixed reference frame $\mathcal{F}_B$.
The rigid body block receives as inputs the total external forces $\mathbf{F}_{\text{total}}$ and moments $\mathbf{M}_{\text{total}}$ acting on the vehicle, expressed in $\mathcal{F}_B$. 
Its outputs include linear velocity $\mathbf{v} = [v_x, v_y, v_z]^\top$  in $\mathcal{F}_B$,
angular rates $\boldsymbol{\omega} = [\omega_x, \omega_y, \omega_z]^\top$ in $\mathcal{F}_B$, vehicle attitude ($\phi$, $\theta$, $\psi$) and position $\mathbf{p}= [p_x, p_y, p_z]^\top$ in $\mathcal{F}_W$ (which is used mostly for visualization).


\subsection{Wing Shape and Segment}
A custom MATLAB function block was developed to parametrize the wing geometry and discretize it into spanwise segments. 
The wing is defined by its total wingspan, root chord and taper ratio, sweep and dihedral angle, and wing rotation angle $\epsilon_P$ for port side and $\epsilon_S$ for starboard side (described in Section \ref{sec:modeling}). The wing rotation angles are defined with respect to hinge axes that are oriented outward from the UAV fuselage (Fig. \ref{fig:UAV_parts_XFLR5}), ensuring a consistent kinematic convention for both wings.
The wing is discretized into multiple spanwise segments (denoted $i$) to account for the non-uniform velocity distribution along the span and the possibility of different airfoils and wing characteristics along the wingspan. Each wing segment is assigned its own local reference frame, denoted as $\mathcal{F}_{P}^{i}$ for the port side or $\mathcal{F}_{S}^{i}$ for the starboard side.
Because the linear velocity varies along the span due to rotational motion, each segment experiences a different velocity $\mathbf{v}_{P}^i$ or $\mathbf{v}_{S}^i$ .

\subsection{Segment Aerodynamics}

For each wing segment $i$, aerodynamic forces and moments are computed independently, where the index $k \in \{P, S\}$ denotes the port or starboard wing side, respectively. 
The intermediate kinematic angle of attack for each segment, ${\alpha_{\text{kin}}}_k^i$, is obtained using the Aerospace Blockset based on the local velocity vectors. This angle is then corrected to account for the commanded wing rotation about the wing axis of rotation ($\epsilon_k$) to determine the effective aerodynamic angle of attack, ${\alpha_{\text{eff}}}_k^i$, according to:
\begin{align}
    {\alpha_{\text{eff}}}_P^i &= {\alpha_{\text{kin}}}_P^i - \epsilon_P,
    \label{eq:aoa_port} \\
    {\alpha_{\text{eff}}}_S^i &= {\alpha_{\text{kin}}}_S^i + \epsilon_S.
    \label{eq:aoa_starboard}
\end{align}

The differing signs account for the antisymmetric rotation kinematics of the opposing wings during the morphing sequence.
Aerodynamic coefficients for each segment are determined using lookup tables parametrized by $Re_{k}^i$ and ${\alpha_{\text{eff}}}_k^i$. For each segment, the lift coefficient $c_l$, drag coefficient $c_d$, and moment coefficient $c_m$ are extracted.

The reference area of each segment $i$ is computed as:
\begin{equation}
A_{k}^{i} = c_{k}^{i} \cdot {b_{\text{seg}}}^{i}_{k},
\end{equation}
where $c$ is the local chord length and $b_{\text{seg}}$ is the span of the segment.
The dynamic pressure for each segment $i$ on wing side $k$ is defined as:
\begin{equation}
q_{k}^{i} = \frac{1}{2} \rho (V_{k}^{i})^2,
\end{equation}
where $\rho$ is the air density and $V_{k}^{i}$ is the magnitude of the air relative velocity at segment $i$ on wing side $k$. Finally, lift and drag forces are computed as \cite{Anderson2016}:
\begin{align}
L_{k}^{i} &= c_l q_{k}^{i} A_{k}^{i}, \\
D_{k}^{i} &= c_d q_{k}^{i} A_{k}^{i}.
\end{align}
The aerodynamic forces are first expressed in the local aerodynamic frame (usually called wind or stability frame):

\begin{equation} \label{aero_forces}
{\mathbf{F}_{aero}}_k^i = \begin{bmatrix} -D_k^i \\ 0 \\ -L_k^i \end{bmatrix},
\end{equation}

and subsequently transformed into the port and starboard frames, $\mathcal{F}_P$ and $\mathcal{F}_S$. From there, they are transformed into $\mathcal{F}_B$ using the kinematic angle of attack ${\alpha_{\text{kin}}}_k^i$ and are denoted as $\mathbf{F}_k^i$ according to:
\begin{equation}
    \mathbf{F}_k^i = 
    \begin{bmatrix}
        \cos{\alpha_{\text{kin}}}_k^i & 0 & -\sin{\alpha_{\text{kin}}}_k^i \\
        0                             & 1 & 0                              \\
        \sin{\alpha_{\text{kin}}}_k^i & 0 & \cos{\alpha_{\text{kin}}}_k^i
    \end{bmatrix}
    {\mathbf{F}_{\text{aero}}}^i_k.
    \label{eq:force_wind_to_body}
\end{equation}

Moments acting on the vehicle arise from two contributions. Firstly, from the moment generated by the aerodynamic force acting at the segment center of pressure, and secondly, from the pitching moment associated with the airfoil moment coefficient.
The moment due to force application is computed as:
\begin{equation}
    {\mathbf{M}_{{arm}}}_k^i = {\mathbf{r}_{AC}}_k^i \times  \mathbf{F}_k^i,
\end{equation}
where ${\mathbf{r}_{AC}}_k^i$ is the position vector of the aerodynamic center acting on the segment, relative to the center of mass, expressed in $\mathcal{F}_B$, and $\mathbf{F}_k^i$ is the segment aerodynamic force.
The aerodynamic pitching moment of the segment is given by:
\begin{equation}
    {\mathbf{M}_{{cm}}}_k^i =
    \begin{bmatrix}
    0 \\
    c_m q_{k}^{i} A_k^i c_k^i \\
    0
    \end{bmatrix},
\end{equation}
where $c_k^i$ is the local chord length.


Selected wing segments are equipped with thrusters. For each thruster-equipped segment $i$ on wing side $k$, the thrust forces are computed from the commanded thrust magnitude $T_k^i$ and the thrust inclination angle $\epsilon_k$, which is corrected again to account for the commanded wing rotation about the wing axis of rotation.
For a thruster-equipped segment $k$, the local thrust force vector is defined as:
\begin{equation}
{\mathbf{F}_{thrust}}_k^i =
\begin{bmatrix}
T_k^i \cos{\epsilon_k} \\
0 \\
T_k^i \sin\epsilon_k
\end{bmatrix}.
\end{equation}
The moment generated by each thrust force about the center of mass is computed using the cross product between the thruster's position vector ${\mathbf{r}_{thrust}}_k^i$ in $\mathcal{F}_B$ and the thrust force vector ${\mathbf{F}_{thrust}}_k^i$:
\begin{equation}
{\mathbf{M}_{thrust}}_k^i = {\mathbf{r}_{thrust}}_k^i \times {\mathbf{F}_{thrust}}_k^i.
\end{equation}

\subsection{Force and Moment Summation}

All external forces are expressed in $\mathcal{F}_B$, like the ones from thrusters and gravity, and moments from the thrusters, and are being summed with the aerodynamic forces and moments to obtain the total loads acting on the vehicle.
Finally, all aerodynamic and propulsion forces and moments from each segment are summed:
\begin{align}
    \mathbf{F}_{total} &=
    \mathbf{F}_g + \sum_{k} \sum_{i}
    \left( \mathbf{F}_{k}^i + {\mathbf{F}_{thrust}}_k^i \right), \\
    \mathbf{M}_{total} &=
    \sum_{k} \sum_{i}
    \left( {\mathbf{M}_{{arm}}}_k^i + {\mathbf{M}_{{cm}}}_k^i + {\mathbf{M}_{thrust}}_k^i \right).
\end{align}

These total forces and moments are then provided as inputs to the 6-DoF rigid-body block, completing the nonlinear dynamic loop.
The resulting model captures the coupled rigid-body dynamics and distributed aerodynamic effects in a fully nonlinear formulation expressed in $\mathcal{F}_B$.

\section{Simulation Analysis}
\label{sec:sim_analysis}

\begin{figure}[t]
\centering
\includegraphics[width=1.0\columnwidth]{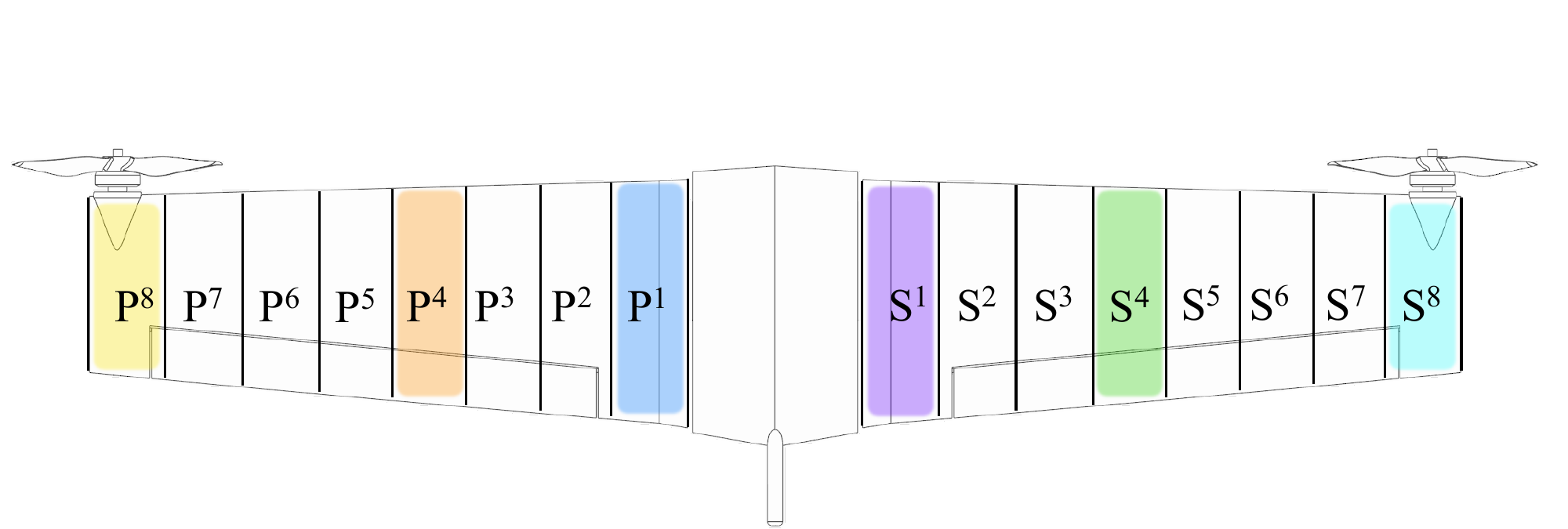}
\caption{Wing segmentation used in the analysis, coloured segments shown according to the simulation graphs.}
\label{fig:seg_color}
\end{figure}

To test the system dynamics in both hover and cruise modes, four simulations are presented to demonstrate how the system behaves in a steady-state trimmed flight and illustrate the system response to step inputs in the wing rotation joints and thruster force. It is important to note that the actuation is open-loop and no real closed-loop control is used for now in these tests. Fig. \ref{fig:seg_color} shows the discretization of each wing into eight segments. The number of segments is determined experimentally, where eight segments per wing were found to be sufficient, as further refinement produced no meaningful change in the results. 

\subsection{Dynamic Take-off Analysis and Hover Stability}

\begin{figure}[t]
\centering
\includegraphics[width=1.0\columnwidth]{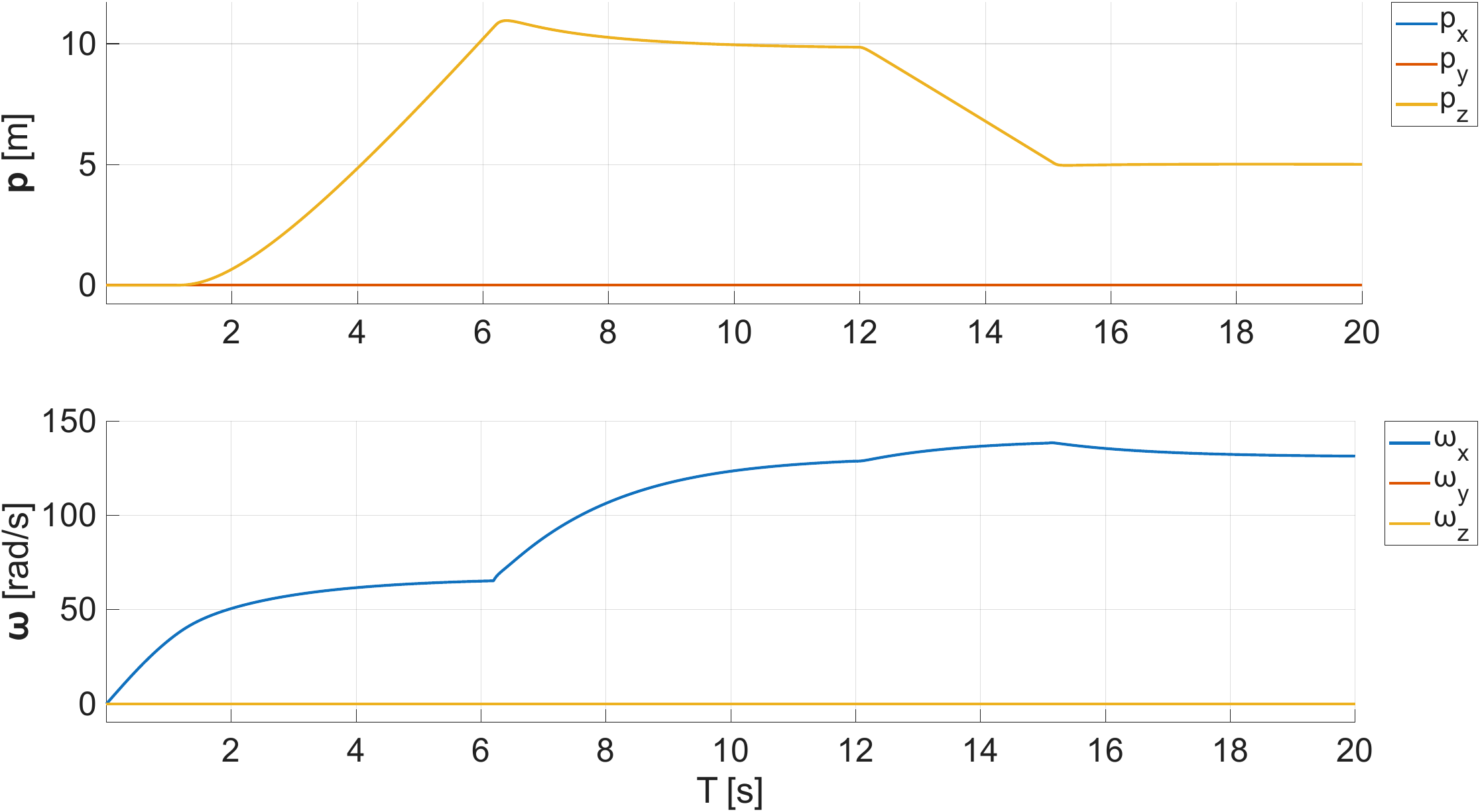}
\caption{Hover mode with actuator control in $\mathcal{F}_B$.}
\label{fig:hov_body_act}
\end{figure}

To test hover mode with actuator inputs, the UAV is initialized with its $x$-axis parallel to the $z$-axis of $\mathcal{F}_W$ at the origin, with gravity acting along $-z$ in $\mathcal{F}_W$. A ground contact model prevents the vehicle from passing through the ground surface. The thrusters are held constant at 0.3 N throughout the experiment, generating a moment about the $x$-axis of $\mathcal{F}_B$ and producing a ground spin-up phase in which $\omega_x$ increases (Fig. \ref{fig:hov_body_act}). As rotational speed increases, the segment air-relative velocities rise and the distributed aerodynamic loads increase until lift-off occurs, after which the vehicle accelerates along the $z$-axis of $\mathcal{F}_W$.

\begin{figure}[t]
\centering
\includegraphics[width=1.0\columnwidth]{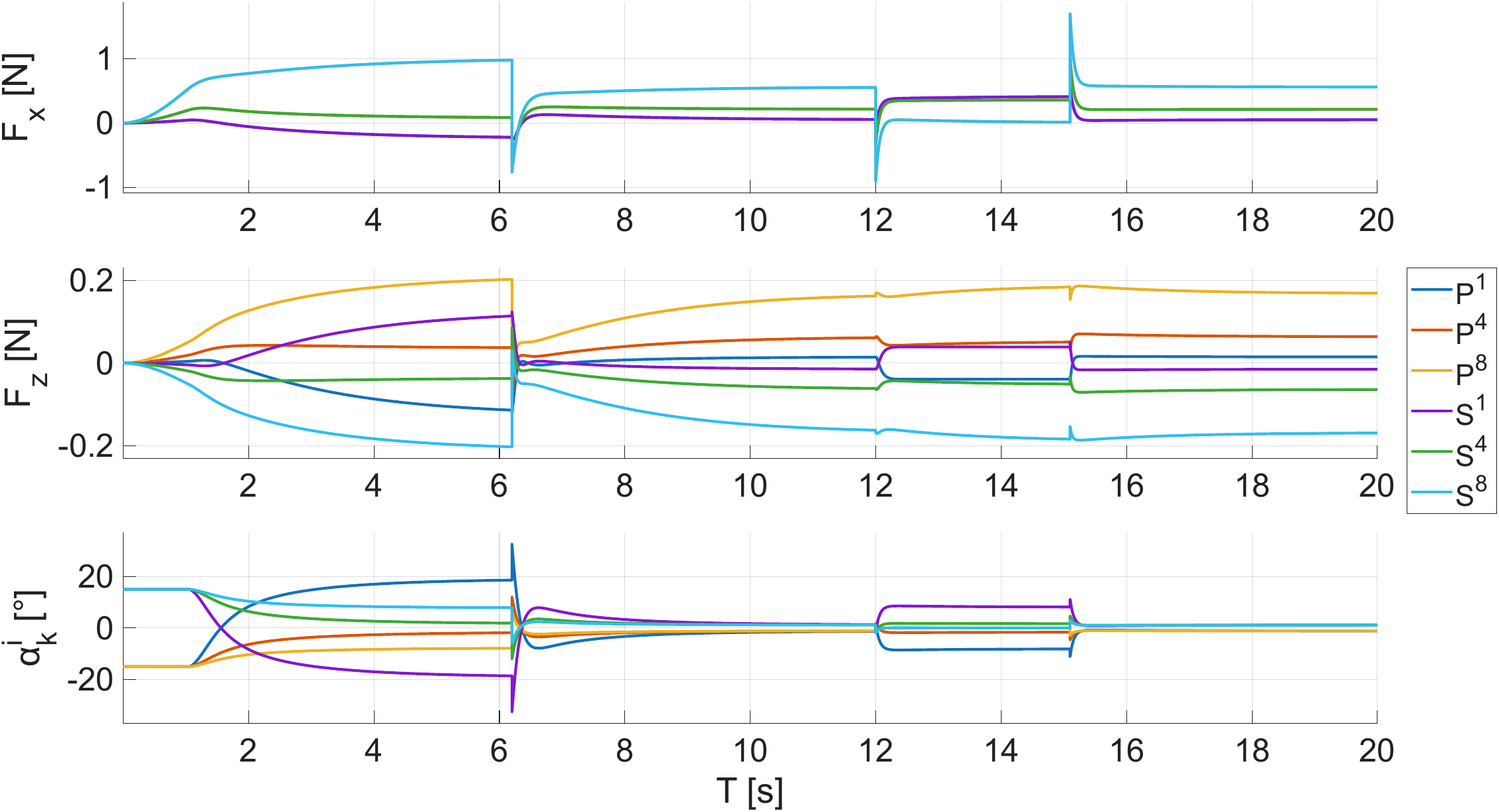}
\caption{Hover mode with actuator control showing the individual segment forces.}
\label{fig:hov_seg_act}
\end{figure}

In Fig. \ref{fig:hov_seg_act}, segment forces are reported in $\mathcal{F}_B$, where $F_x$ is treated as a lift-dominated component and $F_z$ as a drag-dominated component. Since drag is defined along each segment's local relative wind, its projection into $\mathcal{F}_B$ generally distributes across multiple axes and may have opposite signs on the port and starboard wings due to symmetry. After lift-off, the segment angle of attack evolves due to the vector sum of the rotational velocity about the center of mass and the translational climb velocity. This is visible between 2 s and 4 s in Fig. \ref{fig:hov_seg_act}, where the inner segments generate less lift than the outer segments. In the extreme case, segments P1 and S1 produce negative lift because they are moving upward faster than their mechanical angle of attack and rotation speed can generate lift.

\begin{figure}[t]
\centering
\includegraphics[width=1.0\columnwidth]{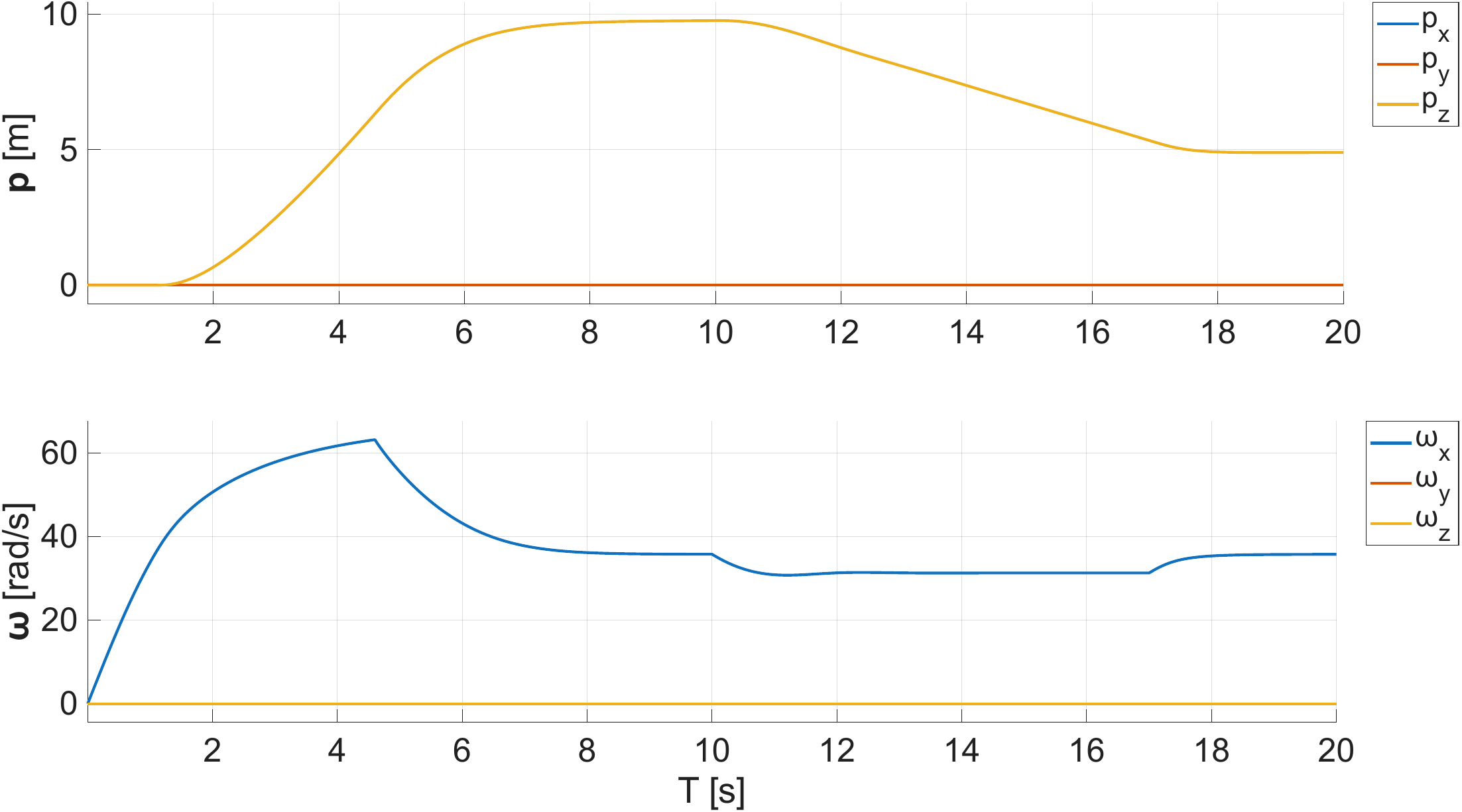}
\caption{Hover mode with thruster control in $\mathcal{F}_B$.}
\label{fig:hov_body_thrust}
\end{figure}

\begin{figure}[t]
\centering
\includegraphics[width=1.0\columnwidth]{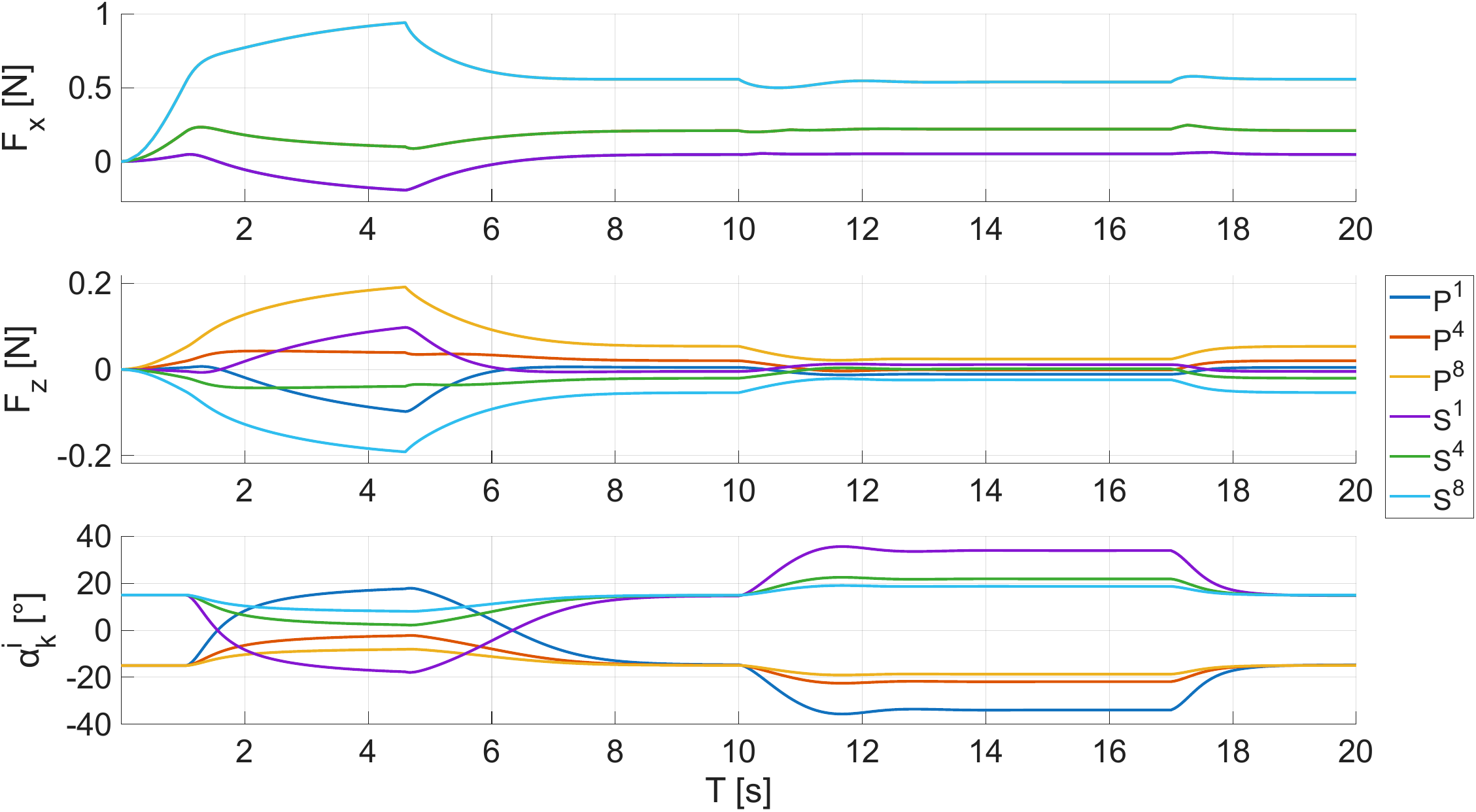}
\caption{Hover mode with thruster control showing the individual segment forces.}
\label{fig:hov_seg_thrust}
\end{figure}

A similar scenario is used to test hover mode with thruster inputs. The joint angles remain fixed at 75$^\circ$, while the thrusters vary in power. A large thrust is applied at the start to take off, a smaller thrust to hover, an even smaller thrust to descend, and then thrust is increased again to hover (Fig. \ref{fig:hov_body_thrust}). The spin rate follows the thruster input. When comparing the outer segments to the inner segments in Fig. \ref{fig:hov_seg_thrust}, the changes in forces and angle of attack follow the same principles as in the actuator test. A reduction in thrust leads to a slower rotation rate, altering the angle of attack distribution across the span, with the inner segments again being highly affected by the translational velocity of the UAV.

\subsection{Morphing-Based Flight Control and Manoeuvrability}
\begin{figure}[t]
\centering
\includegraphics[width=1.0\columnwidth]{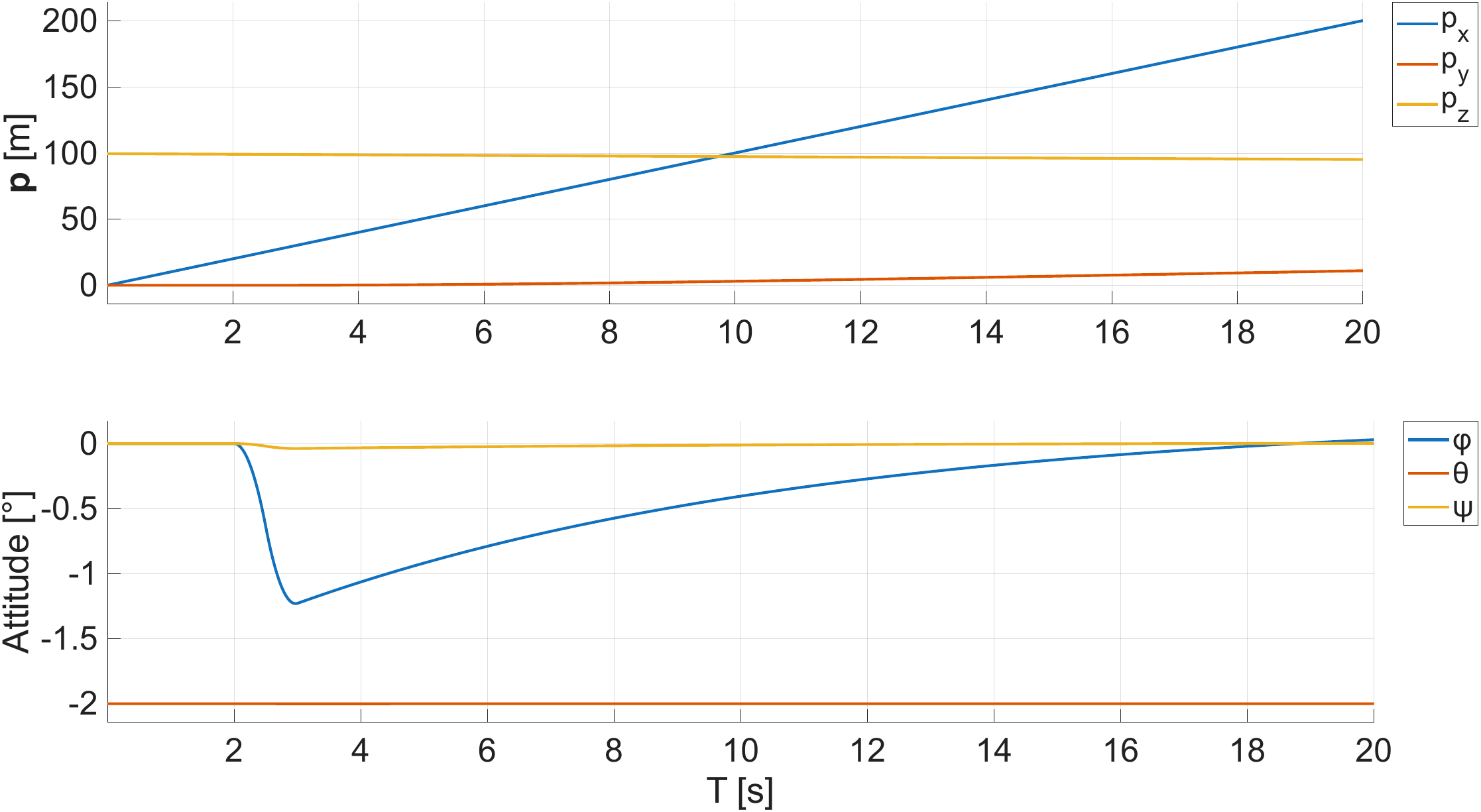}
\caption{Cruise mode changing the roll angle with joint actuators in $\mathcal{F}_B$.}
\label{fig:cruise_body_roll}
\end{figure}

\begin{figure}[t]
\centering
\includegraphics[width=1.0\columnwidth]{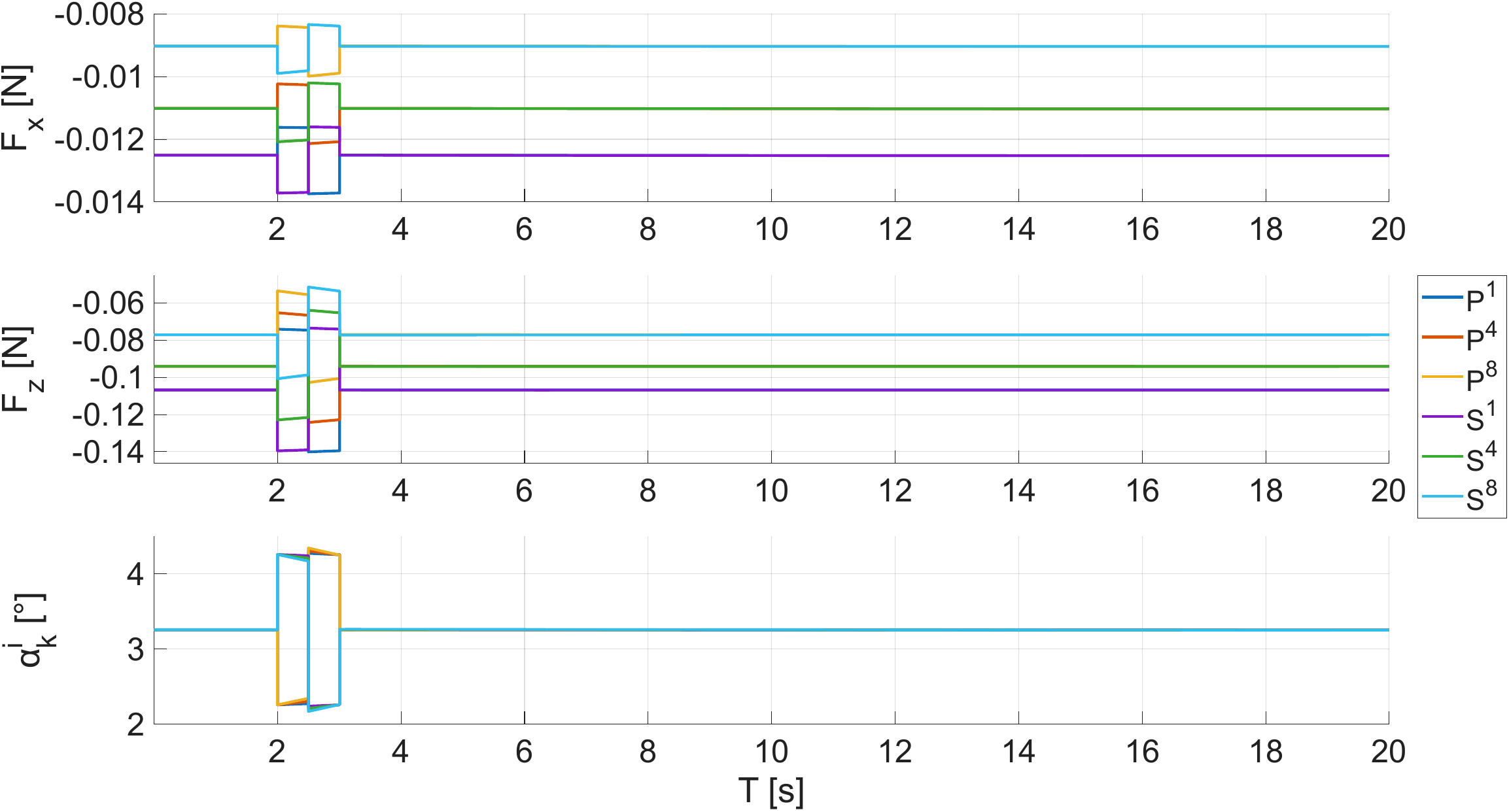}
\caption{Cruise mode changing the roll angle with joint actuators shown on individual segments.}
\label{fig:cruise_seg_roll}
\end{figure}

\FloatBarrier
To evaluate the UAV dynamics in the flying wing configuration with actuator control, a steady glide is simulated. The UAV is initialized with its $x$-axis parallel to the $x$-axis of $\mathcal{F}_W$, and its $z$-axis pointing in the negative $z$-direction of $\mathcal{F}_W$, aligned with gravity. Roll control authority is evaluated by applying opposite $1^\circ$ step inputs to the wing joints, producing a vehicle-level roll response (Fig.~\ref{fig:cruise_body_roll}). At the segment level, this asymmetric joint actuation increases the local angle of attack on the starboard wing and decreases it on the port wing. This generates a differential lift distribution across the segments and between the wings, producing the required moment about the $x$-axis in $\mathcal{F}_B$ (Fig.~\ref{fig:cruise_seg_roll}). When the joints are returned to $0^\circ$, the inherent roll stability of the flying wing is evident in Fig.~\ref{fig:cruise_body_roll}, as the open-loop system naturally damps the roll rate and stabilizes over the subsequent time span \cite{Shams2021}.

\begin{figure}[t]
\centering
\includegraphics[width=1.0\columnwidth]{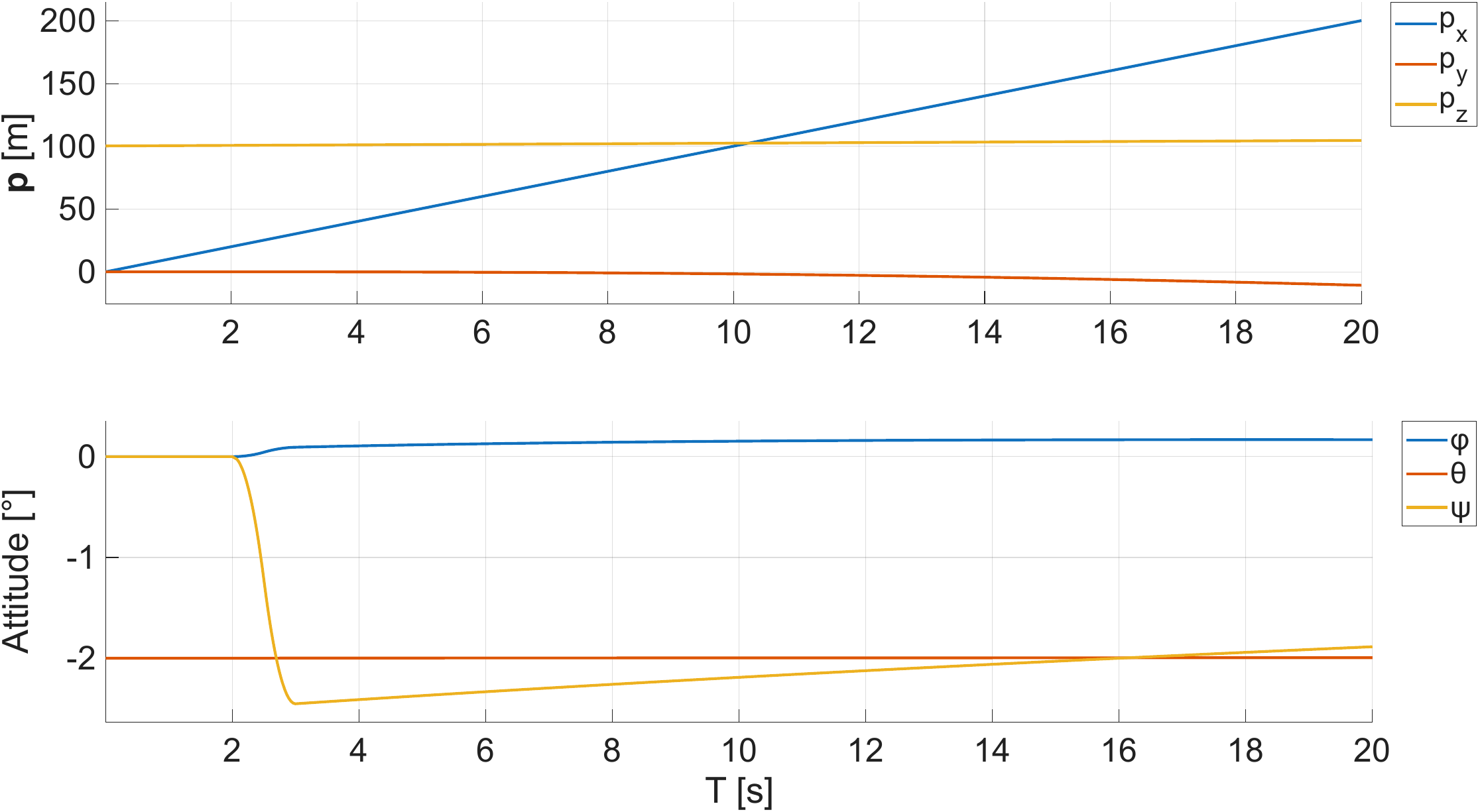}
\caption{Cruise mode changing the yaw angle with thrusters in $\mathcal{F}_B$.}
\label{fig:cruise_body_yaw}
\end{figure}

\begin{figure}[t]
\centering
\includegraphics[width=1.0\columnwidth]{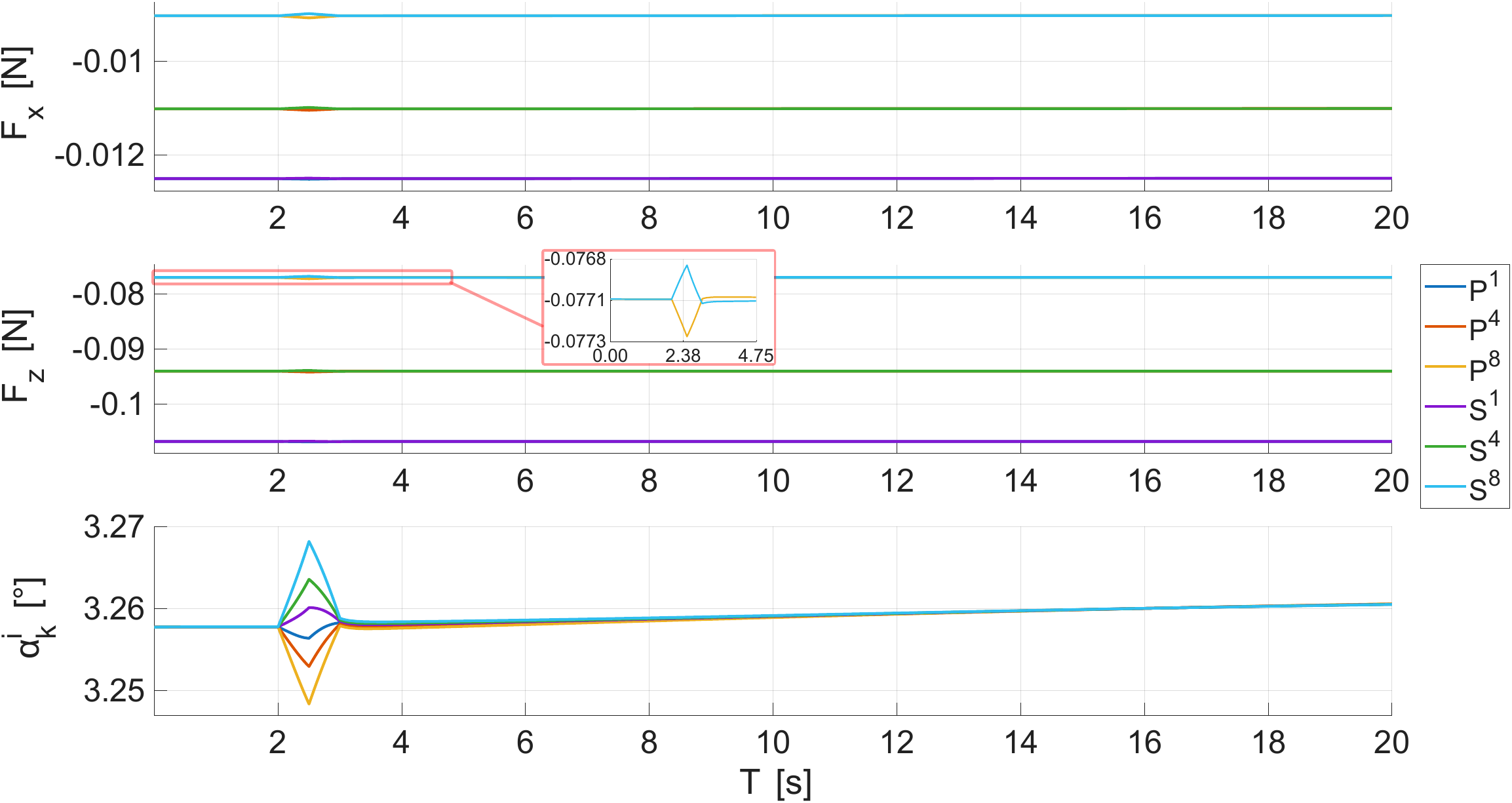}
\caption{Cruise mode changing the yaw angle with thrusters shown on individual segments.}
\label{fig:cruise_seg_yaw}
\end{figure}

A final simulation evaluates the cruise configuration under differential thruster control. The initial vehicle orientation, gravity reference, and trimmed glide state are identical to those in the previous experiment. A yawing moment is introduced about the $z$-axis in $\mathcal{F}_B$ by increasing the port-side thrust by 20\% and decreasing the starboard-side thrust by 20\%, resulting in a negative yaw rate $\omega_z$ (Fig.~\ref{fig:cruise_body_yaw}). Because the flying wing design lacks a vertical rudder and, more importantly in these experiments, any sweep or twist, it does not possess inherent yaw stability \cite{Shams2021}; thus, the yaw angle does not naturally return to zero without active closed-loop control. Furthermore, this induced yaw rate causes the port wing to move faster through the air than the starboard wing, generating an asymmetric lift distribution across the span (Fig.~\ref{fig:cruise_seg_yaw}). This differential lift creates a secondary moment about the $x$-axis in $\mathcal{F}_B$, demonstrating the inherent roll-yaw coupling of the airframe as a positive roll effect emerges in the vehicle-level response (Fig.~\ref{fig:cruise_body_yaw}).

\section{Conclusion and future work}
\label{sec:conclusion}

In this paper, we present a comprehensive nonlinear modeling framework and conceptual design for the MetaMorpher, a metamorphic UAV capable of both rotary-wing hovering and fixed-wing cruise flight. Building on the spincopter platform and incorporating a segmented wing architecture, we address the limitations of traditional rigid-body approximations.
Simulation results confirmed the stability of the spin-up phase in hover mode and the predictable response of individual segments to changes in thrust and wing-folding geometry. Our open-source framework facilitates rapid design iteration for the UAV community.

Future work will focus on experimental validation of the proposed model using a physical MetaMorpher prototype, refinement of aerodynamic parameter identification, thruster modeling as a BLDC motor with propellers, control surfaces modeling and the development of advanced control strategies for mode transition and adaptive morphing.

\appendices
\section*{Acknowledgements}
\footnotesize{This work has been supported by Croatian Science Foundation under the project A High Performance Metamorphic Unmanned Aerial Vehicle - MetaMorpher HRZZ–IP-2024-05-9849 \cite{MetaMorpherweb}.
The work of doctoral student Anja Bosak has been supported in part by the “Young researchers’ career development project--training of doctoral students” of the Croatian Science Foundation.}

\bibliographystyle{ieeetr}
\typeout{}
\balance
\bibliography{Bibliography/bibliography}

@inproceedings{Bai2019,
  title={Design and take-off flight of a samara-inspired revolving-wing robot},
  author={Bai, Songnan and Low, Jun En and Bi, Guilin and Soh, Gim Song and Foong, Shaohui},
  booktitle={2019 IEEE International Conference on Robotics and Biomimetics (ROBIO)},
  pages={1325--1330},
  year={2019},
  organization={IEEE},
  doi={10.1109/ROBIO49542.2019.8961395}
}

@article{Bai2022Splitflyer,
  title={SplitFlyer Air: A Modular Quadcopter that Disassembles into Two Bicopters Mid-Air},
  author={Bai, Songnan and Chirarattananon, Pakpong},
  journal={IEEE/ASME Transactions on Mechatronics},
  volume={27},
  number={6},
  pages={4729--4740},
  year={2022},
  publisher={IEEE},
  doi={10.1109/TMECH.2022.3168434}
}

@article{Crandall1995,
  title={The effect of damping on the stability of gyroscopic pendulums},
  author={Crandall, Stephen H},
  journal={Zeitschrift f{\"u}r angewandte Mathematik und Physik ZAMP},
  volume={46},
  number={1},
  pages={S761--S780},
  year={1995},
  publisher={Springer},
  doi={10.1007/BF00918454}
}

@inproceedings{Low2017,
  author    = {Low, Kin Huat and Win, Aung},
  title     = {Design and Dynamic Analysis of a Transformable HOvering Rotorcraft (THOR)},
  booktitle = {2017 IEEE International Conference on Robotics and Automation (ICRA)},
  year      = {2017},
  pages     = {5405--5412},
  doi       = {10.1109/ICRA.2017.7989755},
  url       = {https://ieeexplore.ieee.org/document/7989755}
}

@inproceedings{Low2018,
  author    = {Low, Kin Huat and Win, Aung},
  title     = {Towards a Stable Three-Mode Transformable HOvering Rotorcraft (THOR)},
  booktitle = {2018 IEEE/ASME International Conference on Advanced Intelligent Mechatronics (AIM)},
  year      = {2018},
  pages     = {---},
  doi       = {10.1109/AIM.2018.8452703},
  url       = {https://ieeexplore.ieee.org/document/8452703}
}

@inproceedings{Orsag2011,
  author    = {Orsag, Matko and Bogdan, Stjepan and Haus, Tomislav and Bunic, Marko and Krnjak, Antonio},
  title     = {Modeling, Simulation and Control of a Spincopter},
  booktitle = {Proceedings of the IEEE International Conference on Robotics and Automation (ICRA)},
  year      = {2011},
  pages     = {2998--3003},
  address   = {Shanghai, China},
}

@inproceedings{Haus2013,
  author    = {Haus, Tomislav and Orsag, Matko and Bogdan, Stjepan},
  title     = {Omnidirectional Vision Based Surveillance with the Spincopter},
  booktitle = {Proceedings of the International Conference on Unmanned Aircraft Systems (ICUAS)},
  year      = {2013},
  pages     = {326--332},
  address   = {Atlanta, GA, USA},
}

@article{Orsag2013,
  author  = {Orsag, Matko and Cesic, Josip and Haus, Tomislav and Bogdan, Stjepan},
  title   = {Spincopter Wing Design and Flight Control},
  journal = {Journal of Intelligent \& Robotic Systems},
  volume  = {70},
  number  = {1--4},
  pages   = {165--179},
  year    = {2013},
  doi     = {10.1007/s10846-012-9725-2},
  url     = {https://doi.org/10.1007/s10846-012-9725-2}
}

@article{Woods2014Adaptive,
  title={The adaptive aspect ratio morphing wing: Design concept and low fidelity skin optimization},
  author={Woods, Benjamin KS and Friswell, Michael I},
  journal={Aerospace Science and Technology},
  volume={36},
  pages={54--64},
  year={2014},
  publisher={Elsevier},
  doi={10.1016/j.ast.2014.03.010}
}

@article{sui2022optimum,
  title={Optimum Design of a Novel Bio-Inspired Bat Robot},
  author={Sui, T. and Zou, T. and Riskin, D.},
  journal={IEEE Robotics and Automation Letters},
  volume={7},
  number={2},
  pages={3419--3426},
  year={2022},
  publisher={IEEE}
}

@article{obradovic2011modeling,
  title={Modeling of flight dynamics of morphing wing aircraft},
  author={Obradovic, B. and Subbarao, K.},
  journal={Journal of Aircraft},
  volume={48},
  number={2},
  pages={391--402},
  year={2011}
}

@article{cevdet2021review,
  title={A review on applications and effects of morphing wing technology on UAVs},
  author={Cevdet, O. Z. E. L. and Ozbek, E. and Ekici, S.},
  journal={International Journal of Aviation Science and Technology},
  volume={1},
  number={01},
  pages={30--40},
  year={2021}
}

@article{li2018review,
  title={A review of modelling and analysis of morphing wings},
  author={Li, D. and Zhao, S. and Da Ronch, A. and Xiang, J. and others},
  journal={Progress in Aerospace Sciences},
  volume={100},
  pages={46--62},
  year={2018},
  publisher={Elsevier}
}

@article{manfreda2018,
  title={On the use of unmanned aerial systems for environmental monitoring},
  author={Manfreda, Salvatore and McCabe, Matthew F. and Miller, Pauline E. and Lucas, Richard and Madrigal, Victor Pajuelo and Mallinis, Giorgos and Ben Dor, Eyal and Helman, David and Estes, Lyndon and Ciraolo, Giuseppe and others},
  journal={Remote Sensing},
  volume={10},
  number={4},
  pages={641},
  year={2018},
  publisher={MDPI},
  doi={10.3390/rs10040641}
}

@article{Anuar2025,
  title={Development of Fixed-Wing VTOL UAVs with the Four-Retractable Rotor Propulsion},
  author={Anuar, Kaspul and Takesue, Naoyuki},
  journal={Journal of Robotics and Mechatronics},
  volume={37},
  number={1},
  pages={510},
  year={2025},
  publisher={Fuji Technology Press Ltd.},
  doi={10.20965/jrm.2025.p0510}
}

@article{Shams2021,
  title={Airfoil Selection Procedure, Wind Tunnel Experimentation and Implementation of 6DOF Modeling on a Flying Wing Micro Aerial Vehicle},
  author={Shams, Taimur Ali and Shah, Syed Irtiza Ali and Javed, Ali and Hamdani, Syed Hossein Raza},
  journal={Micromachines},
  volume={11},
  number={6},
  pages={553},
  year={2020},
  publisher={MDPI},
  doi={10.3390/mi11060553}
}

@article{Coban2023,
  title={Innovative Morphing UAV Design and Manufacture},
  author={{\c{C}}oban, Sezer and Oktay, Tu{\u{g}}rul},
  journal={Journal of Aviation},
  volume={7},
  number={2},
  pages={184--189},
  year={2023},
  publisher={Edit Publishing},
  doi={10.30518/jav.1253901}
}

@inproceedings{Yang2021,
  title={Design and Analysis of a Variable-sweep Morphing Wing for UAV Based on a Parallelogram Mechanism},
  author={Yang, Guang and Guo, Hongwei and Xiao, Hong and Bai, Yue and Liu, Rongqiang},
  booktitle={2021 IEEE International Conference on Robotics and Biomimetics (ROBIO)},
  pages={1018--1023},
  year={2021},
  organization={IEEE},
  doi={10.1109/ROBIO54168.2021.9739420}
}

@article{Yuksek2016,
  title={Transition Flight Modeling of a Fixed-Wing VTOL UAV},
  author={Yuksek, B. and Vuruskan, A. and Ozdemir, U. and Yukselen, M. A. and Inalhan, G.},
  journal={Journal of Intelligent \& Robotic Systems},
  volume={84},
  pages={181--205},
  year={2016},
  publisher={Springer}
}

@article{Win2021,
  title={Design and control of the first foldable single-actuator rotary wing micro aerial vehicle},
  author={Win, Shane Kyi Hla and others},
  journal={Bioinspiration \& Biomimetics},
  volume={16},
  number={6},
  pages={066019},
  year={2021},
  publisher={IOP Publishing}
}

@article{Sufiyan2021,
  title={Joint Mechanical Design and Flight Control Optimization of a Nature-Inspired Unmanned Aerial Vehicle via Collaborative Co-Evolution},
  author={Sufiyan, Danial and Win, Luke Soe Thura and Win, Shane Kyi Hla and Soh, Gim Song and Foong, Shaohui},
  journal={IEEE Robotics and Automation Letters},
  volume={6},
  number={2},
  pages={2044--2051},
  year={2021},
  publisher={IEEE},
  doi={10.1109/LRA.2021.3061373}
}

@article{Sufiyan2022,
  title={An Efficient Multimodal Nature-Inspired Unmanned Aerial Vehicle Capable of Agile Maneuvers},
  author={Sufiyan, Danial and Win, Luke Soe Thura and Win, Shane Kyi Hla and Pheh, Ying Hong and Soh, Gim Song and Foong, Shaohui},
  journal={Advanced Intelligent Systems},
  volume={4},
  number={1},
  pages={2100113},
  year={2022},
  publisher={Wiley Online Library},
  doi={10.1002/aisy.202100113}
}

@article{Zhao2021,
  title={Kriging Aerodynamic Modeling and Multi-Objective Control Allocation for Flying Wing UAVs With Morphing Trailing-Edge},
  author={Zhao, Xinhui and Yang, Yanping and Ma, Xiaoping},
  journal={IEEE Access},
  volume={9},
  pages={59999--60012},
  year={2021},
  publisher={IEEE}
}

@manual{xflr5,
  title = {XFLR5: Analysis of foils and wings operating at low Reynolds numbers},
  author = {Deperrois, Andr{\'e}},
  year = {2019},
  url = {http://www.xflr5.tech/xflr5.htm},
  note = {Open-source aerodynamic analysis tool}
}

@misc{MetaMorpherweb,
  title        = {MetaMorpher project},
  author       = {{UNIZG-FER} and {LARICS}},
  howpublished          ={\url{https://larics.fer.hr/larics/scientific_projects/matamorpher}},
  note         = {Accessed: 2026-02-25}
}

@book{Anderson2016,
  title={Fundamentals of Aerodynamics},
  author={Anderson, John D.},
  edition={6th},
  year={2016},
  publisher={McGraw-Hill Education},
  address={New York, NY},
  isbn={978-1259129919}
}

\end{document}